\documentclass[10pt,twocolumn,letterpaper]{article}

\usepackage[pagenumbers]{cvpr} 

\usepackage{xcolor}
\usepackage{makecell}
\usepackage{tikz}
\usepackage{graphicx}
\usepackage{subcaption}
\usepackage{booktabs}
\usepackage{fancyvrb}
\usepackage[most]{tcolorbox}

\newtcolorbox{promptbox}{
    colback=gray!5,
    colframe=gray!45,
    boxrule=0.4pt,
    arc=2pt,
    left=6pt,
    right=6pt,
    top=6pt,
    bottom=6pt,
    fontupper=\small,
    breakable
}

\usetikzlibrary{tikzmark,calc}

\definecolor{cvprblue}{rgb}{0.21,0.49,0.74}
\usepackage[pagebackref,breaklinks,colorlinks,allcolors=cvprblue]{hyperref}

\def\confName{CVPR}
\def\confYear{2027}
\def\oursName{MemCrafter}

\title{From Seeing to Remembering: Brain-Informed Agentic Text-to-Video \\ Generation for Enhancing Memorability}

\author{Shengqi Dang$^{1,2}$, Yi He$^1$,  Jiaying Lei$^{1,2}$, Feilin Han$^{1,2}$, Nan Cao$^{1,2}$\thanks{Corresponding author.}\\
\\
$^{1}$Tongji University\,\,\,\,\,\,$^{2}$Shanghai Innovation Institute\\
}

\begin{document}

\maketitle

\begin{abstract}
Generating videos that faithfully convey intended
semantics while remaining memorable to viewers is
valuable for applications such as advertising,
education, and storytelling.
However, memorability-oriented text-to-video generation
remains underexplored, as human memory involves
complex neural processes that are difficult to model
and translate into generation guidance. 
In this paper, we propose \oursName{},  brain-informed approach that enhances video memorability through iteratively agentic prompt refinement.
The agentic framework contains two key components.
First, we construct a cortical region-based
neuro-memorability signal that predicts video
memorability from estimated neural responses,
providing both an optimization objective and
interpretable region-level information.
Second, we develop a sensitivity-aware feedback
module, motivated by a conceptual sensitivity
analysis of the optimization objective, to derive
actionable guidance for prompt refinement
while preserving semantic fidelity.
Extensive experiments demonstrate that \oursName{}
consistently improves predicted video memorability
while maintaining semantic alignment and
perceptual quality.
\end{abstract}
    
\section{Introduction}
\label{sec:intro}

Video has become a dominant medium for visual
communication, with streaming and social media
platforms enabling the widespread creation and
dissemination of video content.
As the volume of video content continues to grow,
making videos memorable to viewers has become
increasingly important for real-world applications, such as advertising,
education, and brand communication
~\citep{si2025long,Kumar_2025_WACV}. 

Previous studies have established visual memorability
as the likelihood that visual content will be
remembered by human observers~\citep{khosla2015understanding,cohendet2019videomem}, and developed
computational models to predict the image or video memorability
~\citep{khosla2015understanding,cohendet2019videomem,
newman2020multimodal}.
Building on these advances, recent research has
explored memorability-oriented visual content generation.
For instance, GANalyze~\citep{goetschalckx2019ganalyze},
CogBlender~\citep{dang2026cogblender}, and
MemCoach~\citep{laiti2026take}
investigated ways to enhance image memorability.
However, these approaches have several limitations.
First, many methods~\citep{goetschalckx2019ganalyze, kyledavidson2022modulating} lack semantic control (e.g., through text), limiting their applicability to real-world scenarios.
Second, memorability enhancement often relies on
black-box prediction, offering limited insight
into why visual modifications improve
memorability.
Third, existing methods~\cite{goetschalckx2019ganalyze,dang2026cogblender} primarily focus on static
images, lacking
control over motion and temporal dynamics,
which is important for video memorability~\citep{dumont2023modular}.

These research limitations motivate us to investigate
\textit{memorability-oriented text-to-video generation}
(i.e., generating videos that are more likely to
be remembered by viewers while preserving the
intended semantics of input prompts).
However, achieving this goal presents several
key challenges.
(1) It is challenging to establish a reliable optimization objective, as human memory involves complex neural processes that are difficult
to capture by computational models and black-box prediction may not fully reflect actual human memory.
(2) It is nontrivial to identify effective visual intervention for memorability enhancement, as the relationship between
visual modifications and memorability is complex.
(3) It is difficult to preserve semantic fidelity
when enhancing memorability, as visual interventions
may alter the original content and intent.

A promising approach to reliable memorability
prediction and effective visual intervention
is to leverage brain responses, which reflect
neural processes underlying visual perception
and memory, and whose functional organization
across cortical regions
~\citep{yeo2011organization,schaefer2018local}
provides interpretable guidance based on
known functions, such as object recognition
~\citep{haxby2001distributed}.
Based on this insight, we introduce \textbf{\oursName{}}, a brain-informed
approach for memorability-oriented text-to-video generation.
Specifically, we leverage TRIBE-v2~\citep{d2026foundation},
a brain foundation model capable of predicting
neural responses to visual stimuli, to establish
the relationship between video-evoked
neural responses and video memorability.
Building on this, we develop an agentic
prompt optimization framework that iteratively
refines text prompts to avoid heavily
video generator retraining.
Specifically, we construct a cortical region-based
\textit{Neuro-Memorability Signal} that provides
reliable video memorability and
regional activation information.
To translate this information into effective
generation guidance, we conduct a conceptual
sensitivity analysis of the optimization objective
and introduce a \textit{Sensitivity-Aware Feedback}
module that provides actionable guidance for
the agent to explore effective visual interventions. Additionally, we employ an independent
review agent to provide
semantic alignment scores and diagnostic
feedback for preserving the original
semantic intent.

Extensive experiments demonstrate that \oursName{}
improves video memorability while maintaining semantic
fidelity and visual quality.
User study further supports these improvements in human memory performance.
These findings highlight the potential of
brain information for extending video generation toward desired cognitive effects.

In summary, our contributions are threefold:
\begin{itemize}

\item We are the first, to the best of our knowledge, to propose the task of
\textit{memorability-oriented text-to-video generation}
and we propose \oursName{}, a brain-informed
approach that enhances video memorability through
iteratively agentic prompt refinement.

\item We develop a cortical region-based
    neuro-memorability signal that leverages
    predicted brain responses to estimate video
    memorability and quantify region-wise
    contributions, providing both reliable signal and interpretable guidance for video generation.

\item We conduct a conceptual sensitivity analysis
of the optimization objective and introduce
sensitivity-aware feedback that translates
three complementary factors
into actionable guidance for prompt optimization.

\end{itemize}

\section{Related Work}

We review three closely related research directions:
visual memorability, brain-guided visual generation,
and agentic text-to-video generation.

\subsection{Visual Memorability}

Visual memorability refers to the likelihood that
visual content will be remembered by human observers.
Early studies established the consistency of
memorability across observers and developed
benchmarks for prediction in both image~\citep{khosla2015understanding} and video modals~\citep{cohendet2019videomem,newman2020multimodal}.

Beyond prediction, researchers have explored
how visual content can be modified to enhance
memorability.
GANalyze~\citep{goetschalckx2019ganalyze}
manipulates generative latent representations
to identify visual attributes associated with
image memorability.
CogBlender~\citep{dang2026cogblender}
enables continuous control over cognitive attributes,
including memorability, during text-to-image generation.
MemCoach~\citep{laiti2026take} provides actionable
visual design suggestions for capturing more
memorable images.
In advertising, Si et al.~\citep{si2025long}
developed Henry model for long-term advertisement
memorability prediction and SEED model for generating
memorable advertising scripts, comprising visual
scene descriptions and dialogues.

While these studies have advanced memorability
prediction and memorability-oriented content
manipulation, directly optimizing memorability
during text-to-video generation remains
underexplored.

\subsection{Brain-Guided Visual Generation}

Brain-guided visual generation has emerged as
a way between neural activity and
visual content synthesis.
Existing research has primarily followed two
directions: reconstructing perceived content
from real neural activity and generating
stimuli toward desired neural responses.

The first direction seeks to reconstruct perceived visual content from brain activity. For instance, Takagi and Nishimoto~\citep{takagi2023high}
explored the use of diffusion models to reconstruct
high-resolution images from fMRI signals.
MindEye~\citep{scotti2023reconstructing} combines contrastive learning with diffusion priors for fMRI-to-image reconstruction. DreamDiffusion~\citep{bai2023dreamdiffusion} leverages pretrained diffusion models to generate images from EEG signals.
The second direction, generating
stimuli toward desired neural responses, uses neural response models or
measured feedback to optimize visual stimuli.
BrainDiVE~\citep{luo2023brain} and
NeuroVolve~\citep{chen2025neurovolve}
generate images targeting specific cortical
responses.
NEvo~\citep{tang2026nevo} extends this paradigm
to video generation through evolutionary
prompt optimization, while
MindPilot~\citep{li2026mindpilot}
employs measured EEG feedback for closed-loop
image optimization.

\begin{figure*}[t]
    \centering
    \includegraphics[width=\linewidth]{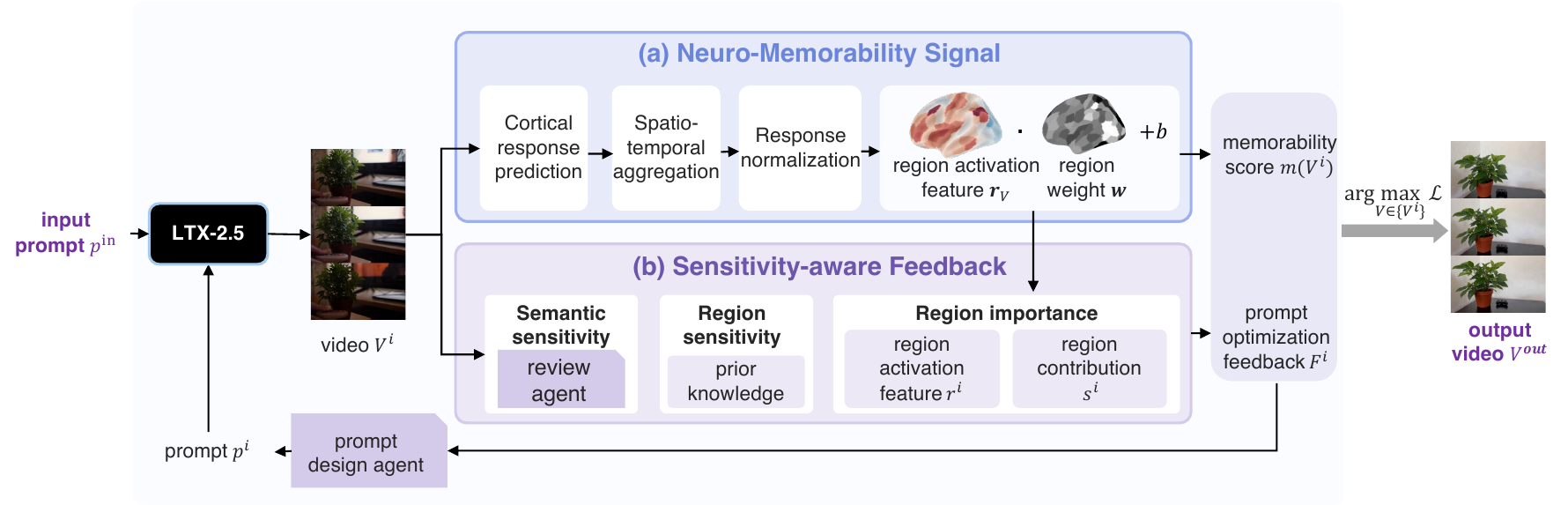}
    \vspace{-0.6cm}

\caption{Overview of \oursName{}.
We employ an agentic approach to iteratively refine
prompts for video generation.}

    \vspace{-0.4cm}
    \label{fig:pipeline}
\end{figure*}

Despite these advances, existing approaches
primarily focus on reconstructing visual content
or optimizing predefined neural response targets.
Our work instead models video memorability
from predicted cortical responses and leverages
region-level feedback to guide generation
toward this cognitive objective.

\subsection{Agentic Text-to-Video Generation}

Agentic text-to-video generation employs
large language model (LLM)-based agents
to iteratively improve video generation
through planning, evaluation, and refinement.
RAPO~\citep{Gao_2025_CVPR} introduces
retrieval-augmented prompt optimization,
while VPO~\citep{cheng2025vpo}
learns prompt refinement from text- and
video-level feedback and preferences.
More recently, agentic approaches have enabled
iterative prompt optimization through
generation, evaluation, and revision.
SCMAPR~\citep{scmapr} combines scenario-aware
prompt rewriting with semantic verification
for complex video generation.
VISTA~\citep{long2026vista} integrates
video evaluation, multimodal critique,
and iterative prompt refinement for
test-time self-improvement.
AgenticGen~\citep{bu2026agenticgen}
further explores reward-guided agentic
video generation for advertising.

These approaches demonstrate the effectiveness
of prompt optimization-driven video refinement, but
primarily focus on semantic alignment,
visual quality, or application-specific objectives.
In contrast, our approach incorporates
memorability feedback into
agentic prompt optimization, enabling
memorability-oriented video generation
while preserving semantic fidelity.

\section{Methodology}
\label{sec:method}

We present \oursName{}, a brain-informed approach for memorability-oriented text-to-video generation.
Given an input prompt $p^{\mathrm{in}}$, our goal is to generate a video $V^{\mathrm{out}}$ with
high memorability while preserving its
intended semantics.
We formulate the generation objective as:
\begin{equation}
    \mathcal{L}(V,p^{\mathrm{in}})
    =
    m(V)+c(V,p^{\mathrm{in}}),
    \label{eq:generation-loss}
\end{equation}
where $m(V)$ denotes the predicted memorability
score of video $V$, and $c(V,p^{\mathrm{in}})$
measures its semantic alignment with the
original prompt.
To preserve the open-domain generalization
capability of pretrained video generators
while avoiding costly model retraining,
we adopt an agentic prompt optimization strategy
for memorability enhancement.
Furthermore, we propose a novel brain-informed
approach that leverages predicted neural signals
not only to provide neurocognitively grounded
memorability predictions, but also to guide
prompt revisions by exploiting the functional
organization of cortical regions
~\citep{yeo2011organization,schaefer2018local}.
For instance, regions associated with visual
perception can inform
targeted prompt modifications to visual appearance~\citep{zeki1991direct}.

As shown in Figure~\ref{fig:pipeline},
our framework employs an iterative prompt optimization
process, where candidate videos are generated and
evaluated to inform subsequent prompt revisions.
Starting from the input prompt $p^1=p^{\mathrm{in}}$, the video
generator $G$ produces a candidate video at
iteration $i$:
\begin{equation}
    V^i=G(p^i),
\end{equation}
where $p^i$ denotes the text prompt at iteration step $i$.
To evaluate each candidate and identify potential
directions for prompt refinement, we introduce
two complementary components.
First, the \textit{Neuro-Memorability Signal}
(Figure~\ref{fig:pipeline}(a)) predicts the
memorability score $m(V^i)$ and provides
interpretable neural information,
revealing how the video activates different
cortical regions and which regional activations
contribute more for memorability.
Second, \textit{Sensitivity-Aware Feedback}
(Figure~\ref{fig:pipeline}(b)) analyzes the
sensitivity of the optimization objective
to prompt modifications and leverages the
identified factors to construct
\textit{prompt optimization feedback} $F^i$.
The prompt design agent $A$ then uses the
memorability score and feedback to update
the prompt:
\begin{equation}
    p^{i+1}
    =
    A(\mathcal{H}^i,m(V^i),F^i),
    \label{eq:prompt-update}
\end{equation}
where $\mathcal{H}^t$ denotes the optimization
history, including the prompts and the evaluation results of
the preceding iterations.
After $N$ iterations, we select the candidate
with the highest optimization objective value as the
final output:
\begin{equation}
    V^{\mathrm{out}}
    =
    \underset{V\in\{V^1,\ldots,V^N\}}{\arg\max}
    \mathcal{L}(V,p^{\mathrm{in}}).
    \label{eq:final-selection}
\end{equation}

We detail the neuro-memorability signal and
sensitivity-aware feedback in the following sections.

\subsection{Neuro-Memorability Signal}
\label{sec:neuro-memorability}

To leverage brain signals for reliable memorability
prediction and effective prompt refinement guidance,
we construct a cortical region-based
neuro-memorability model.
Specifically, we model memorability as a linear
combination of regional activation features to
explicitly quantify each region's contribution:
\begin{equation}
    m(V)=\boldsymbol{w}^{\top}\boldsymbol{r}_V+b,
    \label{eq:mem-signal}
\end{equation}
where $\boldsymbol{r}_V\in\mathbb{R}^{n}$
denotes the regional activation feature vector
of video $V$, with $r_{V,j}$ representing the
activation feature of cortical region $j$.
$\boldsymbol{w}\in\mathbb{R}^{n}$ and $b$
denote the learned region weights and bias,
respectively.
Below we describe 
\textit{Feature Construction}, (i.e., how we construct the feature
\(\boldsymbol{r}_V\)); and \textit{Signal Fitting} (i.e., how
we fit the parameters \((\boldsymbol{w}, b)\)).

\vspace{0.2cm}
\noindent\textbf{Feature Construction.}
To construct regional activation features
$\boldsymbol{r}_V$ that capture cortical
activation patterns while reducing sensitivity
to video duration and temporal response
fluctuations, we perform three steps:
cortical response prediction, spatiotemporal
aggregation, and response normalization,
as illustrated in Figure~\ref{fig:pipeline}(a).

\textit{Cortical response prediction.}
We first employ a pretrained brain foundation model
$\mathcal{E}$ (TRIBE-v2)~\cite{d2026foundation}
to predict cortical responses to video $V$
without requiring actual brain recordings:
\begin{equation}
    \boldsymbol{h}_V=\mathcal{E}(V),
    \label{eq:cortical-prediction}
\end{equation}
where $\boldsymbol{h}_V\in\mathbb{R}^{N_v\times T}$
denotes the predicted functional Magnetic Resonance Imaging (fMRI) responses,
with $N_v$ cortical vertices and $T$ temporal
samples.

\textit{Spatiotemporal aggregation.}
We then aggregate the predicted cortical responses
across space and time within each region to obtain
fixed-dimensional features that accommodate varying
video durations and reduce the influence of
temporal response fluctuations:
\begin{equation}
    \overline{r}_{V,j}
    =
    \frac{1}{|M_j|T}
    \sum_{k\in M_j}
    \sum_{\tau=1}^{T}
    h_{V,k,\tau},
\end{equation}
where $M_j$ denotes the vertex mask of region $j$,
and $h_{V,k,\tau}$ is the predicted cortical response
at vertex $k$ and time $\tau$.

\textit{Response normalization.}
To ensure comparable feature scales across
regions, we standardize the aggregated
responses using training-set statistics:
\begin{equation}
    r_{V,j}
    =
    \frac{\overline{r}_{V,j}-\mu_j}{\sigma_j},
\end{equation}
where $\mu_j$ and $\sigma_j$ are the training-set
mean and standard deviation for region $j$.
The resulting regional activation features
$\boldsymbol{r}_V=[r_{V,1},\ldots,r_{V,n}]^\top$
are used to predict memorability through
Eq.~\eqref{eq:mem-signal}.

\vspace{0.2cm}
\noindent\textbf{Signal Fitting.}
To establish the predictive relationship between
regional activation features and video memorability,
we use VideoMem~\cite{cohendet2019videomem} dataset,
which provides human-annotated memorability scores, and adopt its short-term memorability annotations due to their
higher inter-observer consistency compared with
long-term memorability.
We fit the linear model in
Eq.~\eqref{eq:mem-signal} using ordinary least
squares (OLS).
To improve prediction reliability, we address
two issues: label imbalance and cortical atlas
selection.
To mitigate label imbalance, we partition
videos into memorability-score bins of width
0.05 and randomly sample up to 200 videos per
bin without replacement.
This yields 1,485 videos, which are subsequently
divided into training, validation, and test sets.
Since different cortical parcellations may
affect the predictive utility of regional
activation features, we compare multiple
atlases at different resolutions:
Yeo-7 and Yeo-17~\cite{yeo2011organization},
Destrieux-75~\cite{destrieux2010automatic},
Glasser-180~\cite{glasser2016multimodal},
and Schaefer-100, Schaefer-200, and
Schaefer-400~\cite{schaefer2018local}.
We fit the models on the training set and
select the atlas based on validation performance.
Schaefer-200 achieves the highest validation
performance and is therefore adopted.

\begin{figure*}[t]
    \centering

    \begin{subfigure}[t]{0.29\textwidth}
        \centering
        \includegraphics[width=\linewidth]
        {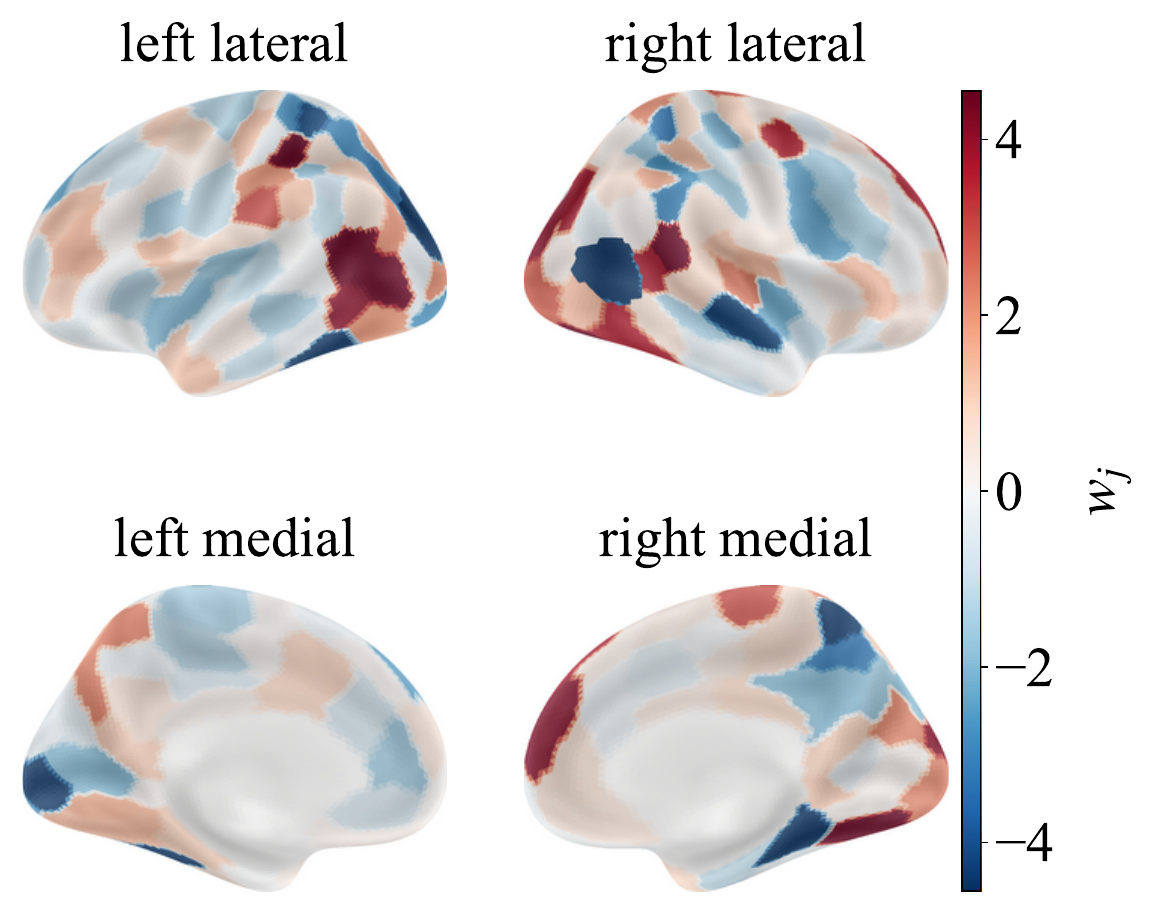}
        \caption{Region-wise importance.}
        \label{fig:signal-weights}
    \end{subfigure}
    \hfill
    \begin{subfigure}[t]{0.38\textwidth}
        \centering
        \includegraphics[width=\linewidth]
        {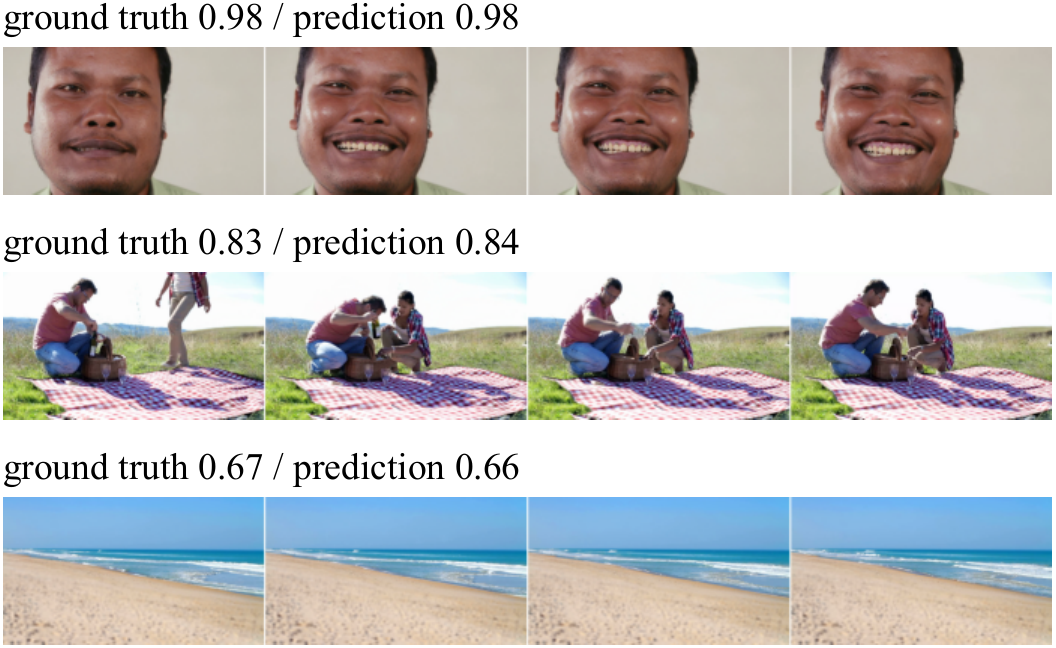}
        \caption{Video samples at different
    memorability levels.}
        \label{fig:signal-examples}
    \end{subfigure}
    \hfill
    \begin{subfigure}[t]{0.26\textwidth}
        \centering
        \includegraphics[width=\linewidth]
        {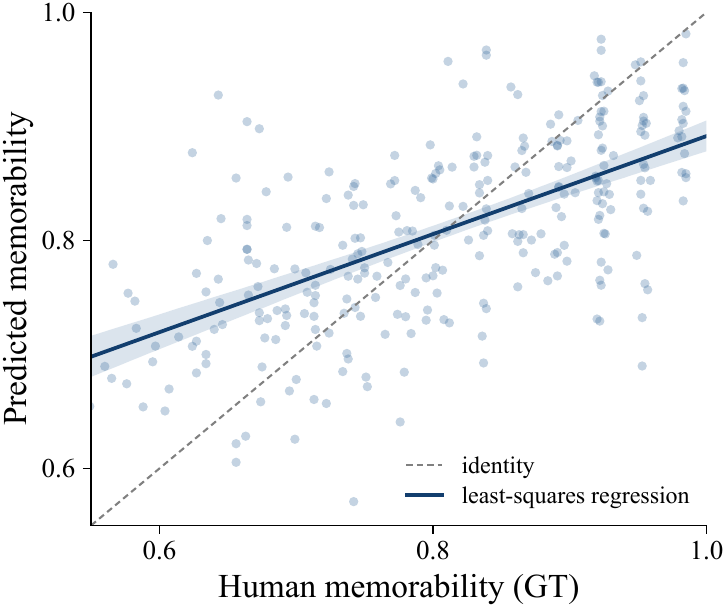}
        \caption{Prediction correlation.}
        \label{fig:signal-correlation}
    \end{subfigure}

    \caption{
    {Evaluation of the proposed neuro-memorability signal.}
    }
    \vspace{-0.3cm}
    \label{fig:signal-evaluation}
\end{figure*}

\begin{figure}
    \centering
    \includegraphics[width=0.8\linewidth]{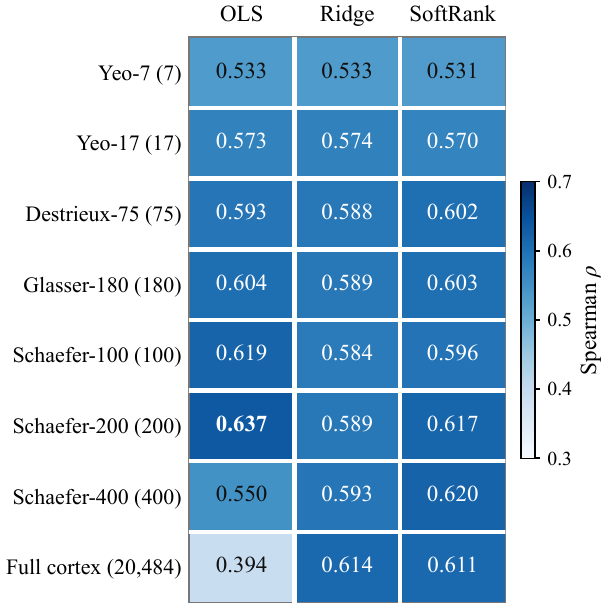}
    \caption{Comparison of different cortical atlases and fitting methods.
The feature dimensions are shown in parentheses.
}
    \label{tab:readout-comparison}
    \vspace{-0.3cm}
\end{figure}

\subsection{Sensitivity-Aware Feedback}
\label{sec:sensitivity-feedback}

Inspired by gradient-based optimization,
we conduct a \textit{conceptual sensitivity analysis}
to identify how prompt modifications may affect
the optimization objective and derive actionable
feedback for iterative prompt refinement.
The sensitivity of the optimization objective
in Eq.~\eqref{eq:generation-loss} to prompt
modifications can be conceptually decomposed as
\begin{equation}
    \frac{\delta\mathcal{L}}{\delta p}
    =
    \sum_{j=1}^{n}
    w_j\frac{\delta r_j}{\delta p}
    +
    \frac{\delta c}{\delta p},
    \label{eq:prompt-sensitivity}
\end{equation}
where $r_j$ denotes the regional activation
feature of the generated video and $c$ denotes
its semantic alignment with the original prompt.
Here, both quantities are understood as functions
of the generation prompt $p$.

As shown in Figure~\ref{fig:pipeline}(b),
this decomposition motivates three complementary
feedback components:
(1) \textit{Region Importance} ($w_j$), which
characterizes the influence of regional activation
features on predicted memorability;
(2) \textit{Region Sensitivity} ($\delta r_j/\delta p$),
which describes how prompt modifications may
affect regional activation features; and
(3) \textit{Semantic Sensitivity} ($\delta c/\delta p$),
which reflects their potential effects on
semantic alignment.
Since prompt-level gradients are unavailable,
we operationalize these factors through distinct
forms of text-based feedback, as detailed below.

\vspace{0.2cm}
\noindent\textbf{Region Importance.}
The learned region weights $\boldsymbol{w}$
in Eq.~\eqref{eq:mem-signal} characterize
the influence of individual regional activation
features on predicted memorability.
However, these weights are fixed across videos
and do not capture video-specific contributions.
We therefore combine them with the regional
activation features
$\boldsymbol{r}^i=\boldsymbol{r}_{V^i}$
of the video generated at iteration step $i$:
\begin{equation}
    s_j^i=w_j r_j^i,
    \label{eq:region-contribution}
\end{equation}
where $s_j^i$ denotes the contribution of region
$j$ to the predicted memorability score of
$V^i$, excluding the bias term.
Both the regional activation features
$\boldsymbol{r}^i$ and region contributions
$\boldsymbol{s}^i=[s_1^i,\ldots,s_n^i]^\top$
are provided to the agent, allowing it to
identify regional activation patterns and
their contributions to predicted memorability.

\vspace{0.2cm}
\noindent\textbf{Region Sensitivity.}
Region importance identifies influential
regional activation features but does not
indicate how they can be modified through
prompt revisions.
To bridge this gap, we leverage neuroscience
knowledge relating cortical functions to
visual processing to inform potential
visual interventions~\citep{corbetta2002control,caspers2010ale,
yeshurun2021default}.
For instance, regions associated with visual
perception may suggest modifications to color,
contrast, and object appearance, whereas
attention-related and sensorimotor regions
may suggest changes in motion and dynamic events.
The agent uses these associations to propose
candidate interventions and evaluates their
effects through subsequent neural feedback.

\vspace{0.2cm}
\noindent\textbf{Semantic Sensitivity.}
Memorability-oriented modifications may cause
generated content to deviate from the original
semantic intent.
To identify such deviations, we employ an
independent review agent.
For each generated video $V^i$, we uniformly
sample eight frames and arrange them into a
$4\times2$ image grid as a visual summary.
Given this summary and the original prompt
$p^{\mathrm{in}}$, the reviewer produces
a semantic alignment score $c^i$ and
diagnostic feedback $f_c^i$.
The diagnostic feedback identifies semantic
deviations, such as missing objects, altered
actions, or inconsistent scenes.
These outputs help the prompt design agent
identify and correct semantic drift during
revisions.

\vspace{0.2cm}
\noindent\textbf{Implementation Details.}
We use GLM-5.3-Flash~\cite{zeng2026glm}
as the prompt optimization agent and
LTX-2.5~\cite{hacohen2026ltx}
as the video generation backbone.
Each video is generated at
$704\times1184$ resolution, 24 fps,
with a duration of 5 seconds.
We perform $N=$20 iteration steps and repeat the experiments with
three random seeds, $\{0,1,2\}$.

\section{Evaluation}
\label{sec:evaluation}

\begin{figure*}
    \centering
    \includegraphics[width=1.0\linewidth]{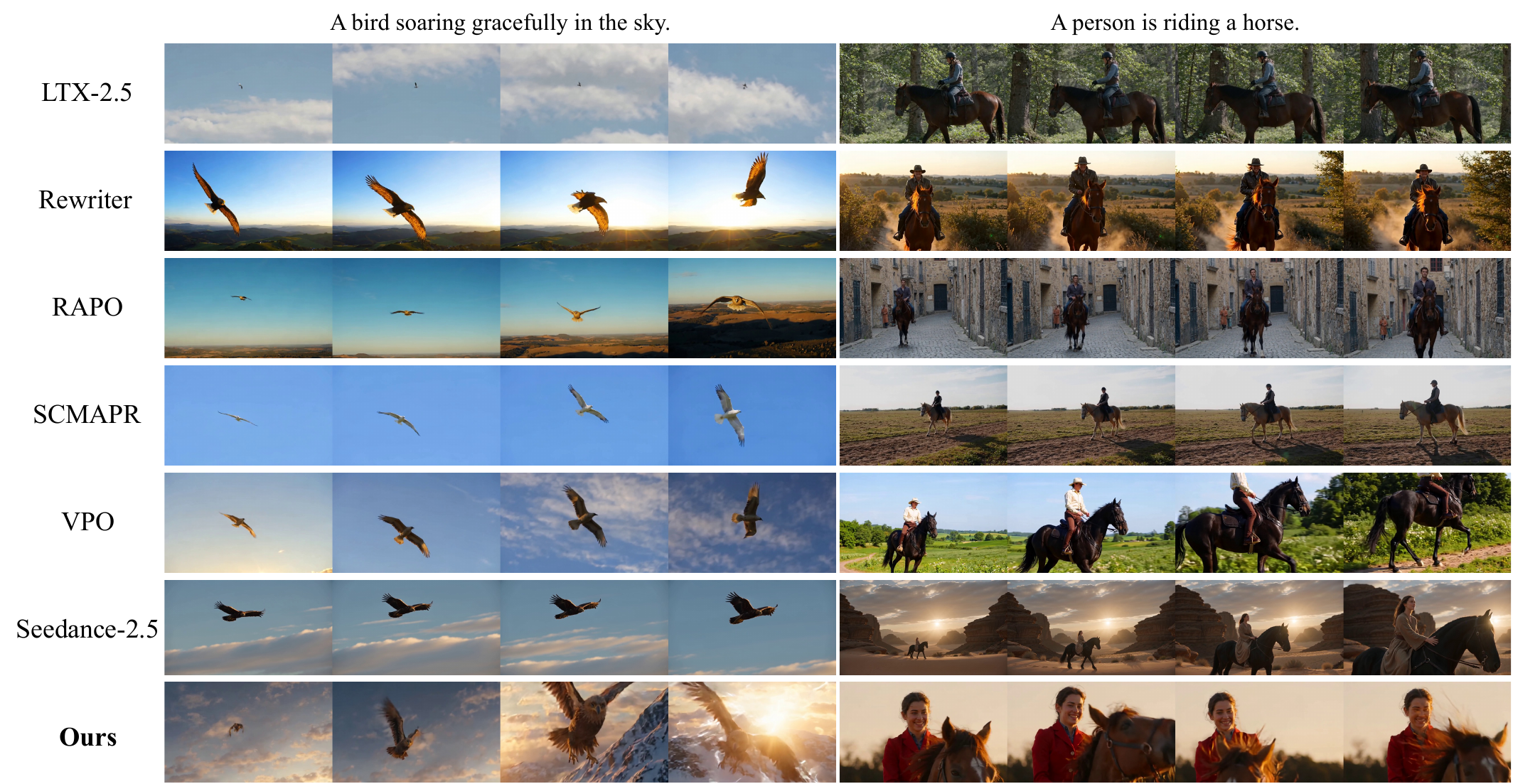}
\caption{
Qualitative comparison.
Our approach enhances video memorability through
effective camera motion (left) and
emphasizing expressive facial features (right),
while preserving the intended semantics.
}
\vspace{-0.4cm}
    \label{fig:qualitative-comparison}
\end{figure*}
\vspace{0.2cm}
We evaluate our framework through four sets of experiments,
each addressing a specific research question:
(1) \textit{Signal Evaluation} (RQ1):
Whether the proposed neuro-memorability signal reliably captures human 
memorability;
(2) \textit{Baseline Comparison} (RQ2):
Whether our framework enhances predicted video memorability
while maintaining semantic alignment and visual quality;
(3) \textit{Ablation Study} (RQ3):
How individual components contribute to memorability enhancement
and semantic alignment; and
(4) \textit{User Study} (RQ4):
Whether the generated videos are more likely to be remembered
by human viewers.

\subsection{Signal Evaluation}
\label{sec:signal-evaluation}

We evaluate the proposed neuro-memorability signal using a test set of 297 videos in VideoMem~\citep{cohendet2019videomem}. We conduct the evaluation from three
aspects: visual analysis, predictive performance,
and cortical atlas and fitting method selection.

\vspace{0.2cm}
\noindent\textbf{Visual analysis.}
Figure~\ref{fig:signal-evaluation}(a) visualizes the
learned region-wise weights $\boldsymbol{w}$
on the cortical surface using the Schaefer-200 atlas.
Positive and negative weights indicate how increases
in regional neural responses affect predicted
memorability.
Figure~\ref{fig:signal-evaluation}(b) presents
representative videos at different memorability
levels, together with their human-annotated and
predicted scores, illustrating the signal's ability
to capture variations in video memorability.
\vspace{0.2cm}
\\
\textbf{Predictive performance.}
Figure~\ref{fig:signal-evaluation}(c) illustrates
the relationship between predicted memorability
and human annotations through a scatter plot
and linear regression fit.
The proposed signal achieves a Spearman correlation
of $\rho=0.637$ on the test set,
indicating a positive association between predicted
and human-annotated memorability.

\vspace{0.2cm}
\noindent\textbf{Cortical Atlas and Fitting Method Selection.}
We systematically investigate the effects of
cortical atlas and fitting method selection.
As shown in Figure~\ref{tab:readout-comparison},
we compare different cortical atlases and
fitting methods (i.e., ordinary least
squares (OLS), ridge regression (Ridge)~\citep{hoerl1970ridge}, and SoftRank~\citep{taylor2008softrank},
a differentiable ranking-based objective).
Under OLS, Schaefer-200 achieves the highest
Spearman correlation $\rho$ (0.637), outperforming
other atlases.
Direct vertex-level
regression yield lower correlations under OLS,
suggesting that region-level aggregation provides
an effective representation for memorability prediction.

\subsection{Baseline Comparison}
\definecolor{highlightorange}{HTML}{E69138}

We compare \oursName{} with baseline methods to evaluate its effectiveness. We first detail the experimental setup, then present quantitative and qualitative comparison,.

\vspace{0.2cm}
\noindent
\textbf{Setup.}
We randomly sample 100 text prompts from the VBench
benchmark~\citep{huang2024vbench} as our test set.

\textit{Metrics.}
We evaluate generated videos using four metrics
(1) \textit{Memorability Enhancement (MS and SMN)},
evaluated using two metrics:
the memorability score (MS), obtained from
our proposed neuro-memorability signal $m(V)$,
and the SemanticMemNet (SMN) score, provided by an independent
video memorability predictor~\citep{newman2020multimodal};
The latter provides an external assessment
of memorability beyond our optimization objective;
(2) \textit{Video Quality (VQ)},
measured by the VBench quality score~\cite{huang2024vbench}, which
aggregates multiple aspects of video quality,
including subject and background consistency
and temporal stability; and
(3) \textit{Semantic Alignment (SA)},
measured by the overall text-to-video consistency
metric in VBench~\cite{huang2024vbench}, which employs
ViCLIP~\citep{wang2024internvid} to evaluate
the text-video semantic alignment.

\newcommand{\sig}[1]{%
  \textsuperscript{\fontsize{5}{6}\selectfont #1}%
}



\newcommand{\res}[3]{%
  \ensuremath{%
    #1%
    ^{\text{\fontsize{5}{6}\selectfont #3}}%
    _{\textcolor{black!70}{\text{\fontsize{5}{6}\selectfont $\pm #2$}}}%
  }%
}

\begin{table}[t]
\centering
\caption{{Baseline comparison.}
Results are reported as mean $\pm$ standard deviation.}
\label{tab:baseline-comparison}

\vspace{-0.15cm}
\small
\setlength{\tabcolsep}{2pt}
\renewcommand{\arraystretch}{1.12}

\begin{tabular*}{\columnwidth}{@{\extracolsep{\fill}}lcccc@{}}
\toprule
\textbf{Method} &
\makecell{MS $\uparrow$} &
\makecell{SMN $\uparrow$} &
\makecell{VQ $\uparrow$} &
\makecell{SA $\uparrow$} \\
\midrule

LTX-2.5 &
\res{0.759}{0.011}{} &
\res{0.847}{0.005}{} &
\res{0.808}{0.008}{} &
\res{0.203}{0.002}{} \\

Rewriter &
\res{0.794}{0.004}{} &
\res{0.849}{0.002}{} &
\res{0.791}{0.007}{} &
\res{0.259}{0.001}{} \\

SCMAPR &
\res{0.773}{0.003}{} &
\res{0.853}{0.005}{} &
\res{0.792}{0.007}{} &
\res{0.244}{0.004}{} \\

RAPO &
\res{0.773}{0.005}{} &
\res{0.857}{0.002}{} &

\res{0.809}{0.005}{} &
\res{0.257}{0.002}{} \\

VPO &
\res{0.790}{0.001}{} &
\res{0.851}{0.001}{} &
\res{0.812}{0.006}{} &
\res{0.244}{0.002}{} \\

Seedance-2.5 &
\res{0.848}{0.003}{} &
\res{0.859}{0.005}{} &
\res{0.815}{0.005}{} &
\res{0.252}{0.004}{} \\

\midrule
\textbf{\oursName{}} &
\res{\textbf{0.908}}{0.003}{} &
\res{\textbf{0.871}}{0.004}{} &
\res{\textbf{0.829}}{0.005}{} &
\res{\textbf{0.261}}{0.006}{} \\

\bottomrule
\end{tabular*}
\vspace{-0.4cm}
\end{table}

\begin{figure*}[t]
    \centering

    \begin{minipage}[t]{0.28\textwidth}
        \centering
        \includegraphics[width=\linewidth]
        {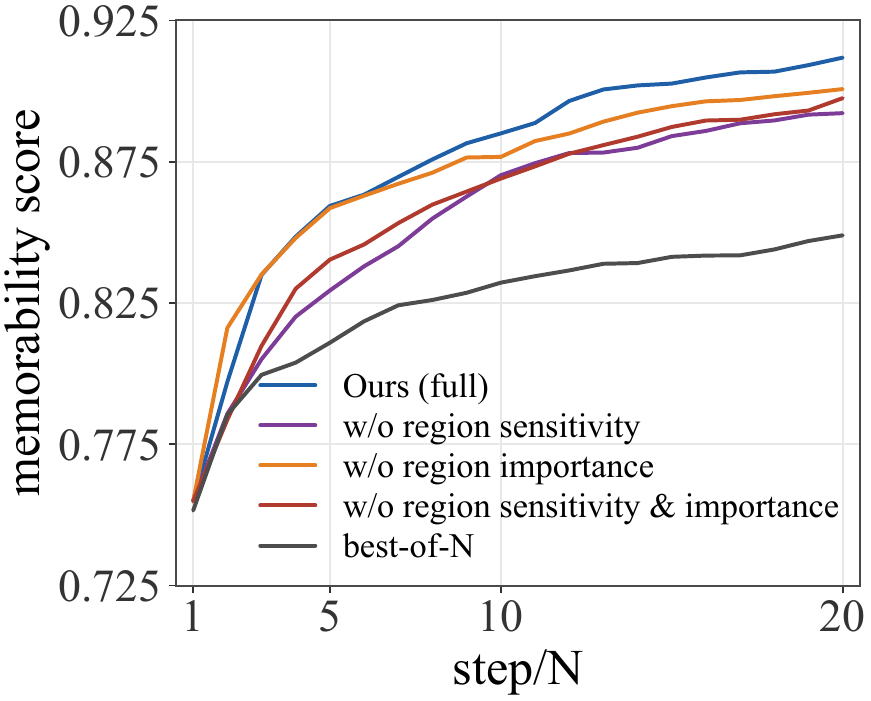}
        \vspace{-0.6cm}
        \captionof{figure}{
        {Results of ablation study.}
        }
        \label{fig:neural-ablation}
    \end{minipage}
    \hfill
    \begin{minipage}[t]{0.62\textwidth}
        \centering
        \includegraphics[width=\linewidth]
        {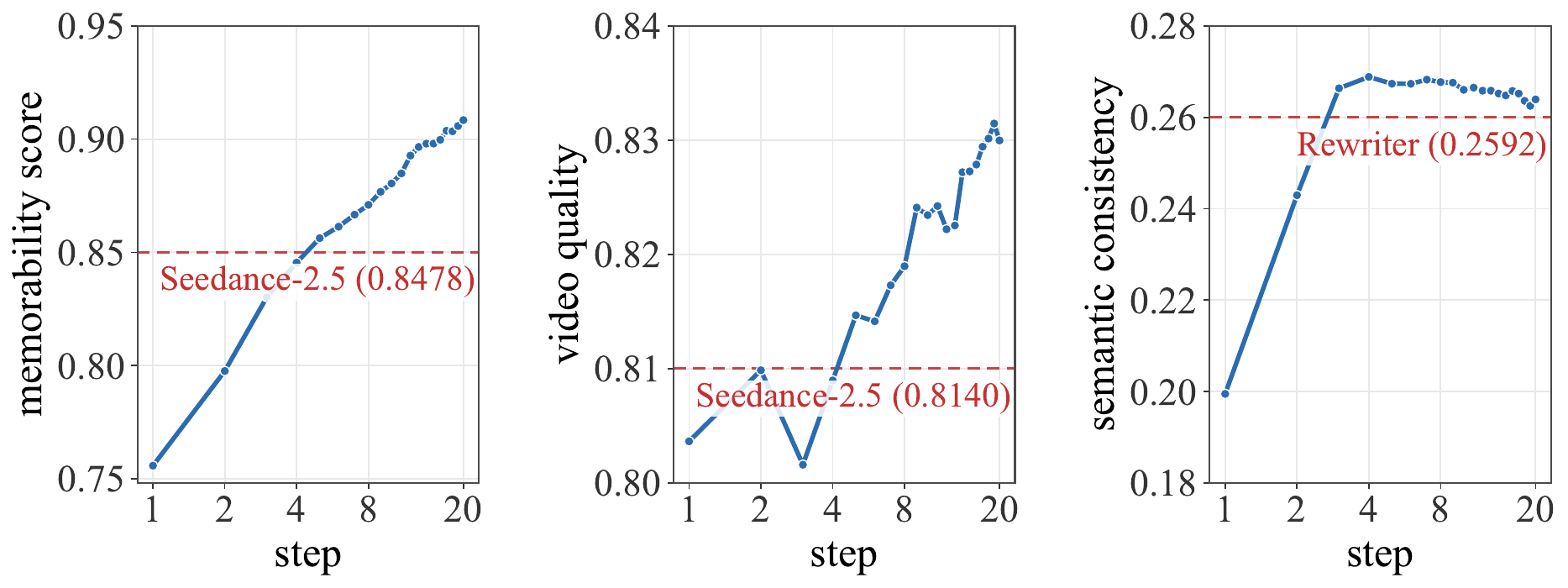}
                \vspace{-0.6cm}
        \captionof{figure}{
        {Test-time scaling analysis. Red dashed lines represent the strongest baseline.}
        }
        \label{fig:test-time-scaling}
    \end{minipage}
\vspace{-0.5cm}
\end{figure*}

\textit{Baselines.}
We compare \oursName{} against six representative
methods:
(1) \textit{LTX-2.5}~\citep{hacohen2026ltx}, which directly generates
videos from the input prompts;
(2) \textit{Rewriter}, employing GLM-5.3-Flash~\citep{zeng2026glm}
to rewrite input prompts for memorability enhancement,
followed by video generation using LTX-2.5;
(3) \textit{RAPO}~\citep{Gao_2025_CVPR}, a retrieval-augmented prompt optimization method; (4) \textit{SCMAPR}~\citep{scmapr}, a self-correcting multi-agent prompt refinement framework for video generation;
(5) \textit{VPO}~\citep{cheng2025vpo}, a prompt optimization method to align video generation with user intent; and (6) \textit{Seedance-2.5}, a state-of-the-art closed-source video generation model, which takes the input prompt augmented with an
instruction to enhance memorability. 

\vspace{0.2cm}
\noindent\textbf{Quantitative Comparison.}
As shown in Table~\ref{tab:baseline-comparison},
\oursName{} achieves the best performance across
all four evaluation metrics.
In terms of memorability, our method achieves
an MS of 0.908, outperforming the strongest
baseline.
Importantly, the improvement also extends to
the independent SMN score, suggesting
that the memorability gains are not confined
to the optimization objective.
Meanwhile, \oursName{} maintains the highest
video quality (VQ) and semantic alignment (SA). 
Although video quality is not explicitly included in our optimization objective, it
improves during optimization,
suggesting that memorability-oriented
refinement can also benefit perceptual quality.

\vspace{0.2cm}
\noindent\textbf{Qualitative Comparison.}
As shown in Figure~\ref{fig:qualitative-comparison},
\oursName{} produces videos with more prominent subjects
and visually distinctive compositions while preserving the semantics of the input prompts.
In the bird-soaring scenario, several baselines depict
the bird as a small subject against a 
static sky, whereas \oursName{} emphasizes the bird
through closer views and dramatic lighting.
Similarly, in the horse-riding scenario, \oursName{}
adopts a closer perspective that highlights the rider's
facial expression.
These examples suggest that brain-informed feedback
encourages visual interventions involving subject
prominence, camera perspective, and lighting,
rather than relying solely on changes in scene appearance.

\subsection{Ablation Study}
We investigate the contributions of iterative prompt
optimization, neural feedback, and
semantic review below.

\vspace{0.2cm}
\noindent\textbf{Effect of Prompt Optimization.}
To assess whether memorability improvements stem
from prompt optimization rather than increased
generation attempts alone, we compare \oursName{}
with a budget-matched variant, Best-of-N.
Specifically, for each iteration step $N$,
Best-of-N independently generates $N$ videos
from the original prompt
and selects the candidate with the highest
memorability score.
As shown in Figure~\ref{fig:neural-ablation},
\oursName{} consistently outperforms Best-of-N
under the same computation budget, demonstrating
the benefit of iterative prompt optimization.

\vspace{0.2cm}
\noindent\textbf{Effect of Neural Feedback.}
To examine the contribution of structured neural
feedback, we remove region importance and region
sensitivity individually and jointly, while
retaining the same iterative optimization procedure.
The variant without both components relies solely
on the scalar memorability score.
As shown in Figure~\ref{fig:neural-ablation},
removing either component reduces memorability
performance, while removing both leads to a
further decline.
These results suggest that region importance
and region sensitivity provide complementary
guidance beyond scalar memorability feedback.

\vspace{0.2cm}
\noindent\textbf{Effect of Semantic Review.}
To evaluate the contribution of semantic review,
we remove the review agent, while retaining the remaining
optimization components.
Without semantic review, the Semantic Alignment
(SA) score decreases from 0.261 to 0.234,
indicating its importance in preserving the
semantic intent.

\begin{figure*}
    \centering
    \includegraphics[width=1.0\linewidth]
    {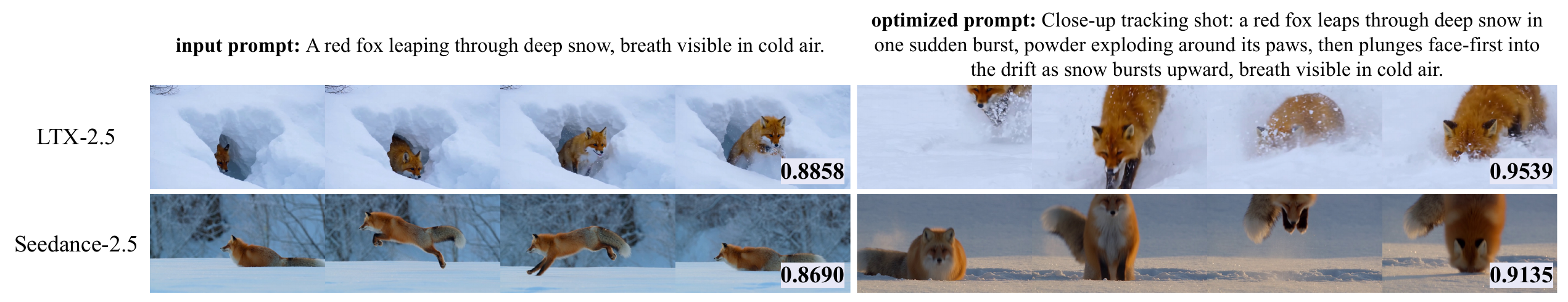}
    \vspace{-0.5cm}
\caption{
Cross-model generalization of optimized prompts.
Prompts optimized using LTX-2.5 also improve
video memorability when transferred to Seedance-2.5. Memorability scores are annotated in the bottom-right corner of each video.
}

\vspace{-0.5cm}
    \label{fig:transfer}
\end{figure*}

\subsection{User Study}
To evaluate whether \oursName{} improves video memorability for human viewers, we conducted a user study following the video recognition paradigm~\citep{cohendet2019videomem}.
We compared \oursName{} with the strongest baseline, Seedance-2.5.

\vspace{0.2cm}
\noindent\textbf{Procedure.}
We randomly selected 20 prompts and generated
one video per method, yielding 40 target videos for evaluation.
We recruited 30 participants, divided into
two groups of 15, to complete the video recognition task,
with each group evaluating 20 target videos
(10 per method).
Participants watched a continuous video stream
and pressed the spacebar whenever they recognized
a previously presented video.
Specifically, each participant viewed 120 videos: 20 target videos presented twice,
60 filler videos presented once, and 10 vigilance videos presented twice to monitor attention.
Target videos reappeared after 45--97 intervening videos to assess video memorability.

\vspace{0.2cm}
\noindent\textbf{Results.}
We quantified video memorability using
the hit ratio, defined for each target video
as the proportion of participants who
correctly recognized it on its second
presentation.
\oursName{} achieved an average hit ratio
of 85.00\%, compared with 77.14\% for
Seedance-2.5, with statistically significant
according to a paired-samples $t$-test
($p<10^{-3}$).
These results indicate that \oursName{}
enhances the memorability of generated videos
for human viewers.

\begin{figure}
    \centering
    \includegraphics[width=1.0\linewidth]
    {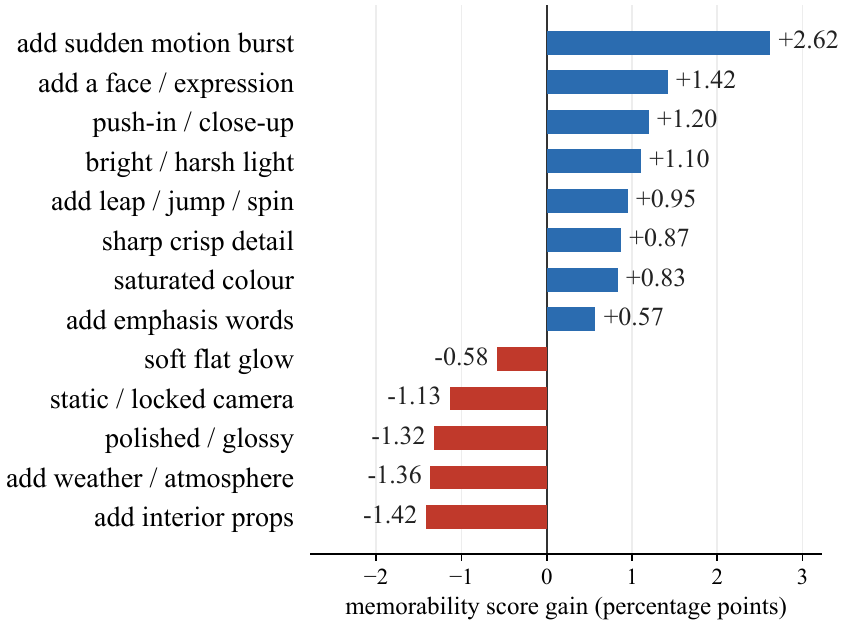}
    \vspace{-0.8cm}
\caption{
Effects of prompt design strategies on video memorability. Blue and red bars indicate positive and negative effects, respectively. The bar lengths represent the effect coefficients (\(\beta\)).
}
\vspace{-0.5cm}
    \label{fig:fig_strategy_clouds}
\end{figure}

\section{Discussion}

\noindent\textbf{Empirical Intervention Strategies.}
Beyond enhancing memorability, our framework provides an
opportunity to investigate the relationship between video
design interventions and cortical response prediction.
As shown in Figure~\ref{fig:fig_strategy_clouds},
we visualize the estimated effect coefficients of
different visual design strategies, which quantify
their associations with changes in predicted
memorability.
We observe that interventions such as sudden motion and close-up shots are
associated with positive memorability effects.
These findings reveal potential associations between
visual design strategies and video memorability.

\vspace{0.2cm}
\noindent\textbf{Test-Time Scaling.}
We further investigate how increasing the number of iteration steps affects performance.
As shown in Figure~\ref{fig:test-time-scaling},
our method achieves better performance than the
strongest baselines across all metrics after 4 steps.
With further optimization, video memorability
and quality continue to improve, while semantic
alignment exhibits a slight decline.
These findings suggest that our framework can benefit from additional test-time computation.

\vspace{0.2cm}
\noindent\textbf{Generalization.}
To investigate cross-backbone generalization, we
directly apply prompts optimized using LTX-2.5
to Seedance 2.5 without additional optimization.
As illustrated in Figure~\ref{fig:transfer},
the optimized prompts also yield noticeable improvements
in predicted memorability on Seedance 2.5,
suggesting the potential transferability of
memorability-oriented design strategies across
different video generation backbones.

\vspace{0.2cm}
\noindent\textbf{Limitations and Future Work.}
Our approach has several limitations.
First, the proposed neuro-memorability signal relies on temporally averaged regional cortical responses and a linear model, without explicitly modeling functional connectivity.
Future work could incorporate functional connectivity to better characterize interactions among cortical regions.
Second, our framework currently focuses on
generating single-shot videos.
Extending it to longer videos and multimodal
content (e.g., audio-visual generation)
introduces additional challenges in modeling
temporal dynamics and cross-modal interactions.
Future work could explore memorability-oriented
generation for more complex and diverse
video scenarios.
Third, our framework relies on iterative test-time prompt optimization, which introduces additional computational overhead.
Future work could distill effective intervention
strategies from optimization trajectories into
reusable agentic skills.

\section{Conclusion}

In this paper, we proposed \oursName{}, a brain-informed
agentic framework for memorability-oriented
text-to-video generation.
Our framework iteratively optimizes generation
prompts for video memorability enhancement.
First, we developed a cortical region-based
neuro-memorability signal that predicts video
memorability from estimated brain responses,
providing both an optimization objective and
interpretable region-level information.
Second, we introduced a sensitivity-aware feedback
module, motivated by a conceptual sensitivity
analysis, to provide guidance for effective prompt refinement.
Experimental results demonstrate that \oursName{}
effectively enhances video memorability while
maintaining semantic alignment and visual quality.
Overall, our work extends text-to-video generation
beyond semantic fidelity toward cognitive effectiveness.
We will release our code to facilitate
reproducibility and future research.

{
    \small
    \bibliographystyle{ieeenat_fullname}
    \bibliography{main}
}


\end{document}